\documentclass[letterpaper, 10 pt, conference]{ieeeconf}  
\IEEEoverridecommandlockouts                              
\usepackage{graphicx}
\usepackage{amsmath} 
\usepackage{amssymb}  
\usepackage[usenames, dvipsnames]{color}
\usepackage[noend]{algpseudocode}
\usepackage{algorithm}
\let\labelindent\relax
\usepackage{enumitem}
\usepackage{ellipsis}
\usepackage{hyperref}

\newcommand{\mydots}{\,\ldots\,}

\newcommand\given[1][]{\:#1\vert\:}

\DeclareMathOperator{\E}{\mathbb{E}}

\title{\LARGE \bf
	Probabilistically Safe Policy Transfer
}

\author{David Held$^{1}$, Zoe McCarthy$^{1}$, Michael Zhang$^{1}$, Fred Shentu$^{1}$, and Pieter Abbeel$^{1,2,3}$ 
	\thanks{$^{1}$
		Department of Electrical Engineering and Computer Science, University
		of California, Berkeley, Berkeley, CA 94720}
	\thanks{$^{2}$International Computer Science Institute (ICSI), University
		of California, Berkeley, Berkeley, CA 94704}
	\thanks{$^{3}$
		OpenAI, San Francisco, CA 94110 
		}
}

\begin{document}
	\maketitle
	\thispagestyle{empty}
	\pagestyle{empty}
	
	
	\begin{abstract}
		Although learning-based methods have great potential for robotics, one concern is that a robot that updates its parameters might cause large amounts of damage before it learns the optimal policy.  We formalize the idea of safe learning in a probabilistic sense by defining an optimization problem: we desire to maximize the expected return while keeping the expected damage below a given safety limit. We study this optimization for the case of a robot manipulator with safety-based torque limits. We would like to ensure that the damage constraint is maintained at every step of the optimization and not just at convergence.  To achieve this aim, we introduce a novel method which predicts how modifying the torque limit, as well as how updating the policy parameters, might affect the robot's safety.  We show through a number of experiments that our approach allows the robot to improve its performance while ensuring that the expected damage constraint is not violated during the learning process.
	\end{abstract}
	
	
	\section{Introduction}
	Reinforcement learning has shown to be a powerful technique leading to impressive performance across a number of domains~\cite{zhong2013value}.  For example, reinforcement learning methods have been used to train a computer to outperform human performance in 49 Atari games~\cite{guo2016deep,mnih2015human}.  Recently, a learning-based method was used to train a computer (AlphaGo) to beat a champion Go player~\cite{silver2016mastering}.  	Similar methods have also been used to train robots to perform a number of difficult tasks  in simulation~\cite{duan2016benchmarking,schulman2015high} and in the real world~\cite{deisenroth2011pilco,peters2008reinforcement}.
	
	However, when applied to robotics tasks in the real world, learning-based methods that adapt their parameters online have the potential to be dangerous. For example, a self-driving car or quadrotor that performs online learning might suddenly adapt its parameters in such a way that causes it to crash into a pedestrian or another obstacle.  
	
	Although there are many ways in which a robot can be dangerous, for this work, we focus on the dangers caused by a robot manipulator applying high torques.  Such a robot might break the object that it is interacting with, break a nearby object, or damage itself.  		
	
	One approach that is often used in practice is to place the robot in an isolated environment while training, where it cannot break anything of importance.  In such an environment, the robot can perform learning with minimal risk.  However, if the isolated training environment is different from the test environment, then this difference can lead to unexpected and possibly dangerous behavior.  For example, suppose we want a robot to operate around or collaboratively with people; the isolated training environment would likely not contain any people, and thus the training environment would not be representative of the test environment in which the robot must operate.  
		
	Another solution is to train the robot in simulation.  Efforts can be made to make the simulation mimic the test environment as much as possible; however, despite one's best efforts, there will likely be differences between simulation and reality, leading to policies that do not work correctly when brought into the real world.  If the robot must re-learn how to behave in the real world, there is a risk of the robot operating dangerously while it is adapting its behavior.
		
	We propose that, when a robot is placed in a new environment, it should initially operate at low torques.   Only once the robot has demonstrated sufficient safety in its new environment do we allow it to operate at higher torques.  Our approach thus makes the assumption that, if the robot violates a safety constraint (which we will define), less damage will be caused if the robot is operating at lower torques.
			
	\begin{figure}[t]
		\begin{center}
			\includegraphics[width=0.95\linewidth]{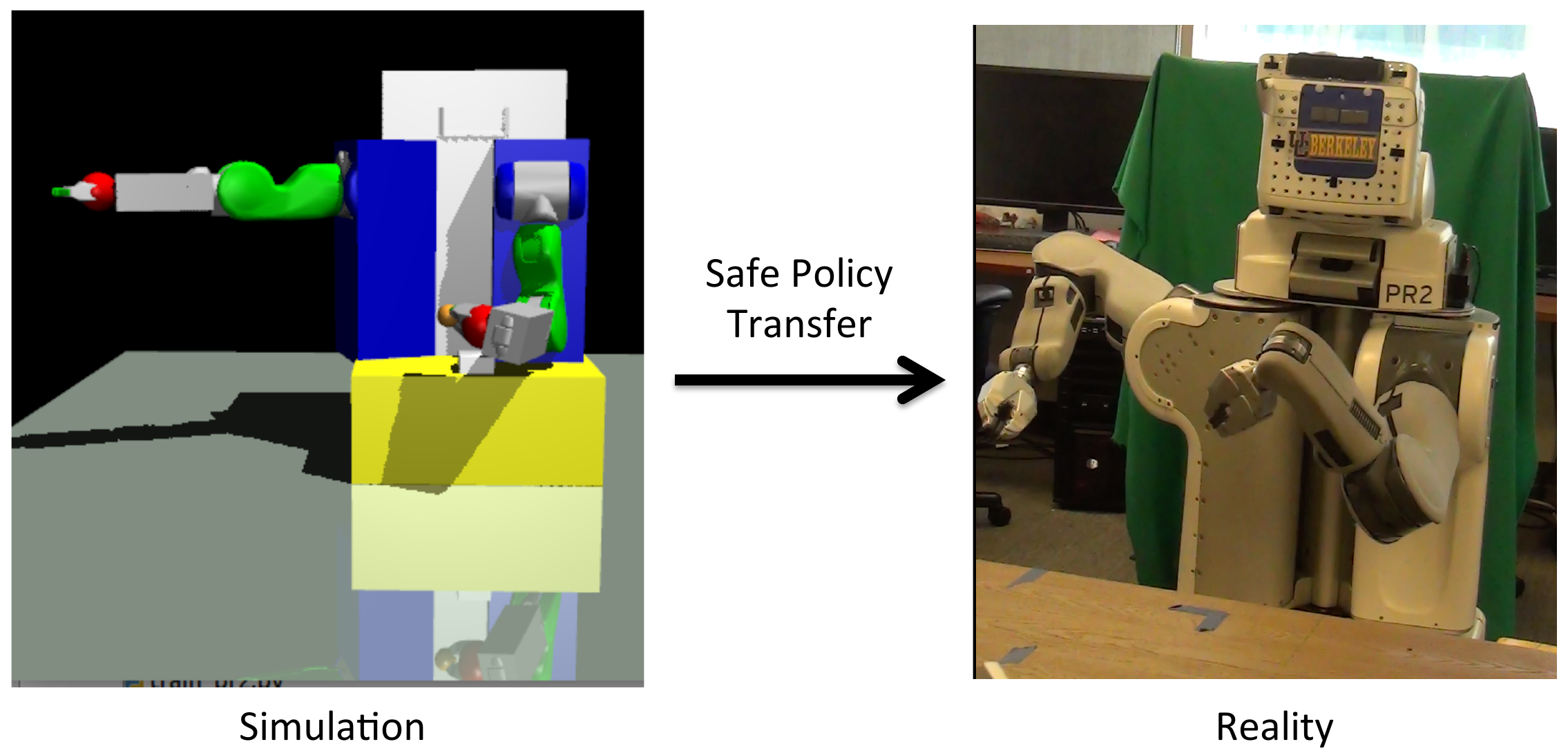}
		\end{center}
		\caption{After training a robot to perform a task in simulation, we adapt the learned policies to the real world.  However, as the robot is adapting, we enforce damage constraints via safety-based torque limits to prevent the robot from applying high torques (and risk breaking something) when the performance is poor.}				
		\label{fig:break}
	\end{figure}
	
	However, imposing safety-based torque limits during training brings up new questions that have not been sufficiently explored in the robot learning literature: How do we define the safety of the robot?  How do we decide whether to increase the torque limits?  How do we ensure that the robot always performs at a safe level of operation even while learning to improve its performance?
		
	We propose an approach to dealing with these challenges for the case of a robot manipulator (shown in Figure~\ref{fig:break}). We evaluate our approach quantitatively, and videos of our results are also available online\footnote{\url{https://youtu.be/fprZHyP\_50o}}.
	We note that our method is not a complete solution to the robot safety problem, and there are still many safety-related challenges that our approach does not address.  Still, we believe our work provides a contribution towards confronting these issues surrounding learning and safe robot operation.
	
	\section{Related Work}
	
	\textbf{Policy Transfer.}
	A number of recent papers have explored how to transfer policies from one domain to another~\cite{cutler2015efficient,cutler2016autonomous,cutler2014reinforcement,fu2015one,mordatchcombining,rusu2016progressive,taylor2011introduction,tessler2016deep}.  
The goal of these papers is typically to minimize the number of samples required in each new environment by transferring knowledge from simulation or from previous environments.   However, these approaches do not investigate the potential safety issues that can occur when a policy is first run in a new environment, before any adaptation has occurred.  
		
	\textbf{AI Safety.}
	Safety has recently become a popular topic within the AI community, as indicated by several recent survey papers on the topic~\cite{amodei2016concrete,garcia2015comprehensive,pecka2014safe}.  Amodei, 
	Olah, et al. organize problems of AI safety into five major categories~\cite{amodei2016concrete}.
	One of their categories is safe exploration: ``how to ensure that exploratory actions in RL agents don't lead to negative or irrecoverable consequences that outweigh the long-term value of exploration."  A related problem is ``robustness to distributional shift$\mydots$how to avoid having ML systems make bad decisions$\mydots$when given inputs that are potentially very different than what was seen during training."  Both of these issues are relevant to the problem discussed in this paper. 
	
	\textbf{Safe exploration.}
	A review of past work on safe exploration is given in~\cite{amodei2016concrete,garcia2015comprehensive,pecka2014safe}.  Our method falls into the category of ``constrained criterion$\mydots$in
	which we want to maximize the expectation of the return while keeping other types of
	expected utilities lower than some given bounds"~\cite{altman1993asymptotic,garcia2015comprehensive}.  Typical constraints involve ensuring that either the expectation of the return exceeds a threshold~\cite{delage2010percentile,geibel2006reinforcement,ponda2013risk}, the variance of the return remains below a threshold~\cite{castro2012policy}, or that any state is reachable from any other state (ergodicity)~\cite{hutter2002self,moldovan2012safe}.  	However, these approaches only guarantee that the constraints will hold at the optimal solution; in contrast to our approach, they do not analyze whether the constraints will hold at every step of the optimization.  
	
	 A recent trajectory optimization algorithm (T-CHOMP) can add a cost that keeps the velocity low near obstacles~\cite{byravan2014space}.  However, this method does not learn to adapt its parameters based on errors in the model.  In contrast, our approach performs online learning while maintaining damage constraints.
		
	\section{Overview}
	In order to obtain safe learning, we pose an optimization problem: 
	we want to train a robot to achieve a task as optimally as possible while maintaining an expected safety constraint.  In order to maintain the expected safety constraint, we impose safety-based torque limits on the robot.  In Section~\ref{sec:Problem Setup}, we will define our optimization problem more precisely, and in Section~\ref{sec:Method}, we will describe our method for maintaining the expected safety constraint while optimizing the policy.
		
	\section{Problem Setup}
	\label{sec:Problem Setup}
		
	\subsection{Minimizing Expected Damage}
	\label{sec:Expected Damage}
	Suppose that a robot behaves according to some policy $\pi(a \given s)$ which maps from states $s$ to actions $a$ (we will abbreviate the policy as $\pi$).  At each timestep $t$, the robot visits a state $s_t$ and takes action $a_t$, leading to a new state $s_{t+1}$ and accrueing a reward, $r_t = r(s_t, a_t, s_{t+1})$.	
	Reinforcement learning problems normally aim to maximize the expected return, defined as the discounted sum of rewards, $R = \sum_t \gamma^t r_t$, for some constant $\gamma$.  
		
	However, operating the robot also has a risk of causing damage to nearby objects or damaging the robot itself.  On any timestep, the agent might cause some amount of damage $d$ to occur.  Our goal is to maximize the expected return while keeping the expected damage below a given threshold:
	\begin{align}
	\label{eq:constrained_opt_gen}
	& \underset{\pi}{\text{maximize}}
	& & \E[R \given \pi] \\
	& \text{subject to}
	\label{eq:constrained_opt2_gen}
	& & \E[d \given \pi] \leq D_{\rm safe}
	\end{align}
	where $\E$ denotes an expectation that incorporates the stochasticity from the environment, the policy, or any other sources of variation.  
	The variable $D_{\rm safe}$ is a constant that specifies how much expected damage we are willing to tolerate.  
	For example, in the autonomous vehicle domain, we might specify that we want the expected number of major accidents per 100 million miles traveled to be less than 0.1.
	Note that the damage $d$ is assumed to be a non-negative quantity (i.e. $\forall \pi, d \geq 0$), so we can set $D_{\rm safe}$ arbitrarily close to 0 to minimize the amount of expected damage that the robot might cause.  We refer to equation~\ref{eq:constrained_opt2_gen} as the damage constraint.
	
	The problem that we wish to deal with is that, when we first operate the robot using our initial policy, the robot might violate the damage constraint.  Further, as we optimize the policy, the policy may change in ways that also violate the constraint.  Our goal is then to limit the amount of expected damage caused by the robot while the policy is being adapted. 
			
	\subsection{Task Safety}
	In order to minimize expected damage, the robot must have some notion of what might cause damage.  We define a safety constraint:
	\begin{align}
	\label{eq:lim_constraint}
	u(s) \leq u_{\rm lim}
	\end{align}
	where $u(s)$ is some function over states.  For example, we can specify that an autonomous vehicle must remain some minimum distance from any pedestrian.  We assume that the robot might cause damage only when the constraint is violated (i.e. when $u(s) > u_{\rm lim}$).  To discourage the robot from violating this constraint, we use a penalty method for optimization, with the reward received at time-step $t$ as
	\begin{align}
	\label{eq:penalty_reward}
	r'_t = r_t - \lambda \max(0, u(s_t) - u'_{\rm lim})^2
	\end{align}
	where $u'_{\rm lim} \leq u_{\rm lim}$.	By increasing $u'_{\rm lim}$ over time until $u'_{\rm lim} = u_{\rm lim}$ and by increasing $\lambda$ over time, we can ensure that our final policy, at convergence, will maximize the expected return without violating the safety constraint.
		
	However, the problem is that penalty methods only guarantee that the safety constraint (equation~\ref{eq:lim_constraint}) will not be violated at convergence.  On the other hand, for robotics, any unsafe action can be dangerous or harmful, including actions taken before convergence.  We thus design a method that ensures that, even if the robot violates the safety constraint at intermediate steps of the optimization, the robot will not cause a significant amount of damage.  
	
	We now define the expected damage in terms of safety violations.  We assume that, due to differences between simulation and reality and due to uncertainty in the environment, avoiding violating the safety constraint entirely is impossible.    
	Let $U$ be the event of violating the safety constraint from equation~\ref{eq:lim_constraint} (and possibly causing damage) when acting under policy $\pi$, and let $p_u(\pi) = p(U)$.  We refer to $p_u(\pi)$ as the ``unsafety rate," which we abbreviate as $p_u$.  Note that no damage can occur unless the event $U$ has occurred.  
	
	Further, let $d_{\rm max}(\pi)$ be the maximum amount of damage that can be caused by policy $\pi$ if it violates the safety constraint (as explained below, different policies can cause different amounts of damage during constraint violations).  Then we can upper-bound the expected damage by 
	\begin{align}
	\label{eq2}
	\E[d \given \pi] &= \int p(d(\pi)) \, d(\pi) \, \textrm{d}d \\
	\label{eq:total_prob}
	&= \int p(U) \, p(d(\pi) \given U) \, d(\pi) \, \textrm{d}d \\
	& \leq \int p(U) \, p(d(\pi) \given U) \, d_{\rm max}(\pi) \, \textrm{d}d \\
	& =  p(U) \, d_{\rm max}(\pi) \int p(d(\pi) \given U) \,  \, \textrm{d}d \\
	& =  p(U) \, d_{\rm max}(\pi)
	\end{align}
	where equation~\ref{eq:total_prob} comes from the total probability theorem combined with the fact that $p(d(\pi) \given \neg U) = 0$ because we assume that damage can only occur when the constraint is violated.
	Thus our damage constraint (equation~\ref{eq:constrained_opt2_gen}) becomes
	\begin{align}
	p_u(\pi)d_{\rm max}(\pi) \leq D_{\rm safe}.
	\end{align}
	Our goal is to impose constraints on the policy such that, even if the policy violates the safety constraint (i.e. even if $p_u(\pi) > 0$), the amount of expected damage will never exceed the threshold $D_{\rm safe}$.
	
	\subsection{Torque Limits}
	\label{sec:Torque Limits}
	To concretize our problem, we assume that the actions chosen by the robot are torques that will be applied to its various motors.
	In order to limit the damage $d_{\rm max}(\pi)$ that can be caused by the robot, we introduce torque limits into our robot operation.  At each time step, we limit the torque (in absolute value) that the robot can output to some threshold $T_{\rm lim}$.  Specifically, if our policy $\pi$ outputs a torque $T$, then we instead apply a torque 
	\begin{align}
	\label{eq:truncation}
	T' = \max(\min(T, T_{\rm lim}), -T_{\rm lim}).
	\end{align}
	We use the same torque limit for all joints.
	
	By applying torque limits, we limit the amount of damage that can be caused by the robot acting under any policy.  For example, if an autonomous vehicle is driving slowly, then even if it moves unsafely (due to, for example, a slippery road or poor visibility conditions), its slow speed will result in less damage if the vehicle gets into an accident.  Similarly, if a robot manipulator bumps into a fragile object, it will have a lower chance of breaking the object if it is moving at lower torques, and if it bumps into a person, then the person will be less likely to become injured.  
			 
	The damage that can be caused by the robot is now a function of both the policy $\pi$ and the torque limit $T_{\rm lim}$, which we write as $d_{\rm max}(\pi, T_{\rm lim})$.  Analyzing the damage that can be caused at different torque levels is outside of the scope of this paper, so we assume that the maximum amount of damage that can be caused in an unsafe state is linear in the torque limit, i.e. 
	\begin{align}
	\label{eq:linear_damage}
	d_{\rm max}(\pi, T_{\rm lim}) = \alpha \, T_{\rm lim}
	\end{align}
	for some constant $\alpha > 0$.  Our method does not depend on the specific form of this function, and other forms can be easily used.  We can fold $\alpha$ into the safety limit $D'_{\rm safe} = D_{\rm safe} / \alpha$, so without loss of generality, we assume that $\alpha = 1$.  
	
	Our damage constraint now becomes
	\begin{align}
	\label{eq:constraint}
	p_u(\pi, T_{\rm lim})  \, T_{\rm lim} \leq D_{\rm safe}
	\end{align}
	where the probability of violating the safety constraint $p_u$ is now a function of both the policy $\pi$ as well as the torque limit $T_{\rm lim}$, since both affect the robot's operation.  This equation is fairly intuitive: if the robot is operating at a low torque limit which is unlikely to cause much damage, we accept a greater probability of violating the safety constraint; if the robot is operating at a larger (and more dangerous) torque limit, then we insist on a lower probability of violating the safety constraint.	Our approach is not restricted to torque limits but can be used with any type of limit that controls the amount of damage that can be caused by an unsafe policy, such as velocity constraints.
			
	Our initial policy might be unsafe, so we initially force the robot to operate at a low torque limit.  Although a robot operating at a lower torque limit can cause less damage, limiting the torques that the robot can apply also might limit the rewards that can be obtained by the robot.  A robot that can only apply low torques will take a longer time to achieve each task, and thus it will typically accrue lower rewards (for many typical reward functions).  We define the Probabilistically Safe Policy Transfer problem as that of deciding how to modify the policy and adjust the torque limit $T_{\rm lim}$ in order to maximize the expected return while maintaining the expected damage constraint.
	
	One potential side-effect of this framework is that, in some cases, varying the torque limits can increase the convergence time of the policy: for each new value of the torque limit, the policy might need to adapt its behavior, leading to slower convergence.  However, we accept this cost in convergence time due to the benefit of safer robot behavior.
	
	\section{Method}
	\label{sec:Method}
	
	\subsection{Overall Algorithm}
	\label{sec:Overall algorithm}
	Now that we have defined what it means for a robot to operate safely (in a probabilistic sense), we describe a method for safely optimizing a policy.  
	The overall algorithm for Probabilistically Safe Policy Transfer is shown in Algorithm~\ref{alg}.  We begin with an initial policy $\pi$ and an initial torque limit of $T_{\rm lim} = T_{\rm min}$ where $T_{\rm min} < D_{\rm safe}$, which ensures that, even if $p_u = 1$, our expected damage $p_u \, T_{\rm lim}$ will not exceed the damage limit.  Using our initial policy and torque limit, we perform a set of rollouts.  Based on these rollouts, we update the policy, as is normally done in reinforcement learning.  
	
	Next, we adjust our torque limit.  We would like to increase our torque limit as much as possible, but we need to ensure that the damage constraint will continue to hold for the next iteration with the updated policy.  We achieve this as follows: we first emperically compute the current unsafety rate $p_u$ from the rollouts.  We then anticipate how much the unsafety rate might increase in the next iteration, given by $\Delta p_u$.  Let $\Delta p_{u1}$ be the amount that the unsafety rate might increase as a result of changing the torque limit (described in Section~\ref{sec:Varying Torque Limit}), and let $\Delta p_{u2}$ be the amount that the unsafety rate might increase as a result of changing our policy (described in Section~\ref{sec:Updating the params}).  
	The probability of the union of these cases is bounded by the sum of the individual probabilities.  Using this bound gives us the worst-case estimate of:
	\begin{align}
	p'_{u} &= p_u + \Delta p_u \\
	\label{eq:pf_adjust}
	&\leq p_u + \Delta p_{u1} + \Delta p_{u2}.
	\end{align}
	Then, we adjust (increase or decrease) the torque limit $T_{\rm lim}$ such that, if the unsafety rate increases to $p'_u$, the damage constraint would hold with equality, i.e. we set 
	\begin{align}
	\label{eq:tl_adjust}
	T_{\rm lim} = D_{\rm safe} / p'_u.
	\end{align}
	We then perform new rollouts using our updated policy and new torque limit, and we repeat until convergence.
	
	\begin{algorithm}
		\caption{Probabilistically Safe Policy Transfer}\label{alg}
		\begin{algorithmic}[1]
			\renewcommand{\algorithmicrequire}{\textbf{Input:}}
			\Require{Initial policy $\pi$, limit $D_{\rm safe}$, initial torque limit $T_{\rm min}$}
			\State{$T_{\rm lim} \gets T_{\rm min}$} 
			\Comment{Initialize the torque limit}
			\While{not converged} 
			\State{$\tau \gets$ Run rollouts of policy $\pi$ with torque limit $T_{\rm lim}$}
			\State{Update the policy $\pi$} 
			\State{Compute safety-related quantities:}
			\State\hspace{\algorithmicindent}{Compute unsafety rate $p_u$ from rollouts $\tau$}	
			\State\hspace{\algorithmicindent}{$p'_{u} \gets p_u + \Delta p_u$} 
			\Comment{Predict the next unsafety rate}
			\State\hspace{\algorithmicindent}{$T_{\rm lim} \gets D_{\rm safe} / p'_u$}
			\Comment {Update the torque limit}
			\EndWhile
		\end{algorithmic}
	\end{algorithm}
	
	\subsection{Changing the Torque Limit}
	\label{sec:Varying Torque Limit}
	
	We assume that our policy maps states to a Gaussian
	distribution with mean $\mu(s)$ and variance $\sigma^2(s)$, and to choose an action, we sample from this distribution.  This is a common representation in deep reinforcement learning for continuous control problems~\cite{levine2016end,mnih2016asynchronous,schulman2015trust}.  
	
\begin{figure}[ht]
		\begin{center}
			\includegraphics[width=0.95\linewidth]{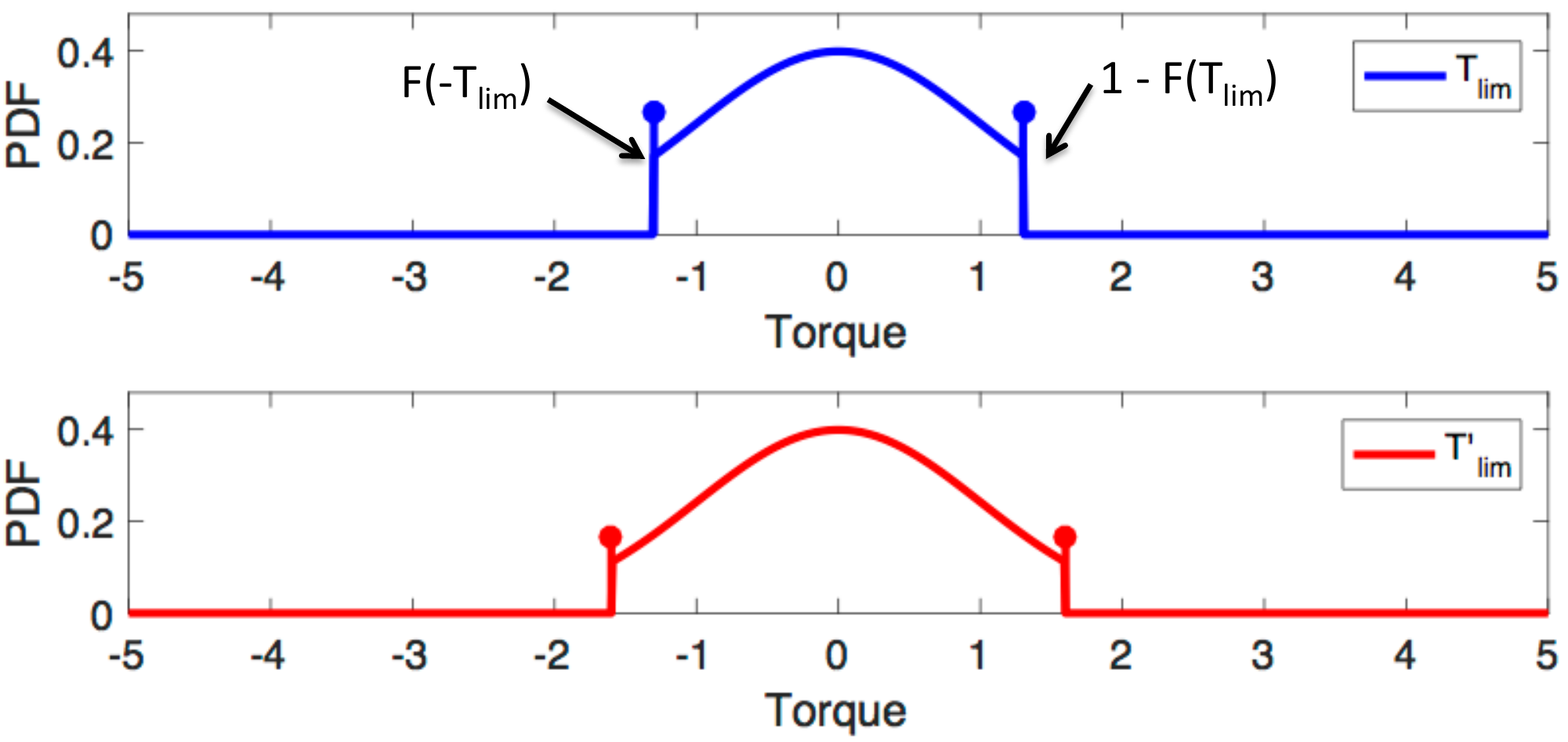}
		\end{center}
		\caption{Top: Probability density function over the old torque limit (after truncation).  There is an additional discrete probability mass at the (positive and negative) torque limits.  Bottom: Probability density function from the new torque limit (after truncation).}
		\label{fig:inc_limit}
	\end{figure}

	To ensure that the damage constraint is not violated, we impose safety-based torque limits on our robot.  The torque $T$ is sampled from the distribution $\mathcal{N}(\mu, \sigma^2)$ and then truncated down to a torque $T'$ based on the torque limit (see Equation~\ref{eq:truncation}).  The distribution for the applied torque (i.e. after truncation) can thus be represented by a truncated Gaussian:
	\begin{align}
	\label{eq:tlim_cases}
	p(T' | s) = 
	\begin{cases}
	\frac{1}{\sigma \sqrt{2 \pi} }\exp\Big(\frac{-(T' - \mu)^2}{2 \sigma^2}\Big), & \text{if } -T_{\rm lim} < T' < T_{\rm lim} \\
	F_{\mu, \sigma}(-T_{\rm lim}), & \text{if } T' = -T_{\rm lim} \\
	1 - F_{\mu, \sigma}(T_{\rm lim}), & \text{if } T' = T_{\rm lim} \\
	0 & \text{otherwise}
	\end{cases}
	\end{align}
	where $F_{\mu, \sigma}(T)$ is the cumulative distribution function of the Gaussian with mean $\mu$ and variance $\sigma^2$ (we have omitted the dependency of the the mean and variance on the state $s$).  Because truncation is applied after sampling, $T_{\rm lim}$ contains the probability mass of all torques $T \geq T_{\rm lim}$; similarly, $-T_{\rm lim}$ contains the probability mass for all torques $T \leq -T_{\rm lim}$.  This distribution is shown in Figure~\ref{fig:inc_limit} (top).
	
	The probability density resulting from an increase in the torque limit from $T_{\rm lim}$ to $T'_{\rm lim}$ is illustrated in Figure~\ref{fig:inc_limit} (bottom).  
	The probability density remains the same for all torques for which $-T_{\rm lim} < T < T_{\rm lim}$, and the probability density changes for torques in the range
	\begin{align}
	\label{eq:tails}
	T_D = (T \leq -T_{\rm lim}) \cup (T \geq T_{\rm lim}).
	\end{align}
	The probability that a torque would previously have been sampled from this range is given by 
	\begin{align}
	\label{eq:prob_t_in_td_s}
	p(T \in T_D | s) = 1 - F_{\mu, \sigma}(T_{\rm lim}) + F_{\mu, \sigma}(-T_{\rm lim}).
	\end{align}
	
	This distribution is conditioned on the state $s$.  In order to compute the unconditional probability over torques, we must integrate over the different states:
	\begin{align}
	p(T \in T_D) = \int_s p(s) \, p(T \in T_D | s) \, ds.
	\end{align}
	The distribution over states $p(s)$ that are visited depends on the actions that are taken; thus, $p(s)$ depends on the policy $\pi$ as well as the torque limit $T_{\rm lim}$.  However, estimating the effect of the torque limit on $p(s)$ is complex, so we approximate $p(s)$ using the distribution from the previous iteration (a similar approximation is often made for policy gradient methods; see, e.g.,~\cite{schulman2015trust}). 
	In practice, we replace this integral with the Monte-Carlo estimate, i.e. 
	\begin{align}
	\label{eq:prob_t_in_td_exp}
	p(T \in T_D) &= \E_{s \sim p(s)} [p(T \in T_D | s)] \\
	\label{eq:prob_t_in_td_exp2}
	&= \E_{s \sim p(s)} [1 - F_{\mu, \sigma}(T_{\rm lim}) + F_{\mu, \sigma}(-T_{\rm lim})].
	\end{align}
	This quantity is the probability that a different torque will be sampled due to changing the torque limit.
	If all such cases lead to an unsafe action by our policy, then by increasing the torque limit, the unsafety rate might increase by 
	\begin{align}
	\Delta p_{u1} = p(T \in T_D).
	\end{align}
	This quantity is used by our algorithm in Equations~\ref{eq:pf_adjust} and~\ref{eq:tl_adjust} to determine the new torque limit.
		
	Note that our estimate of $\Delta p_{u1}$ does not depend on the amount by which the torque limit was increased.  Our approach is fairly conservative: we assume that if any torque $T$ is sampled that is greater than our previous torque limit, then this new torque might lead to an unsafe action, regardless of how close that torque is to the previous torque limit.  Although such an approach might in some cases be  conservative, our method errs on the side of ensuring safe robot operation.
	
	The above analysis holds only for torque limit increases and not decreases.  Nonetheless, in the case of torque limit decreases, we still use the above formula to predict changes to the unsafety rate.  Based on the above analysis, such an approach is conservative, because decreasing the torque limit (i.e. making the torque limit tighter) will cause no new torques to be sampled that could not have been sampled on the previous iteration.  Thus, using the same formula in both cases results in a method that is both simpler and more conservative than handling each case separately.
				 
	\subsection{Updating the policy} 
	\label{sec:Updating the params}
		
\begin{figure}[ht]
		\begin{center}
			\includegraphics[width=0.5\linewidth]{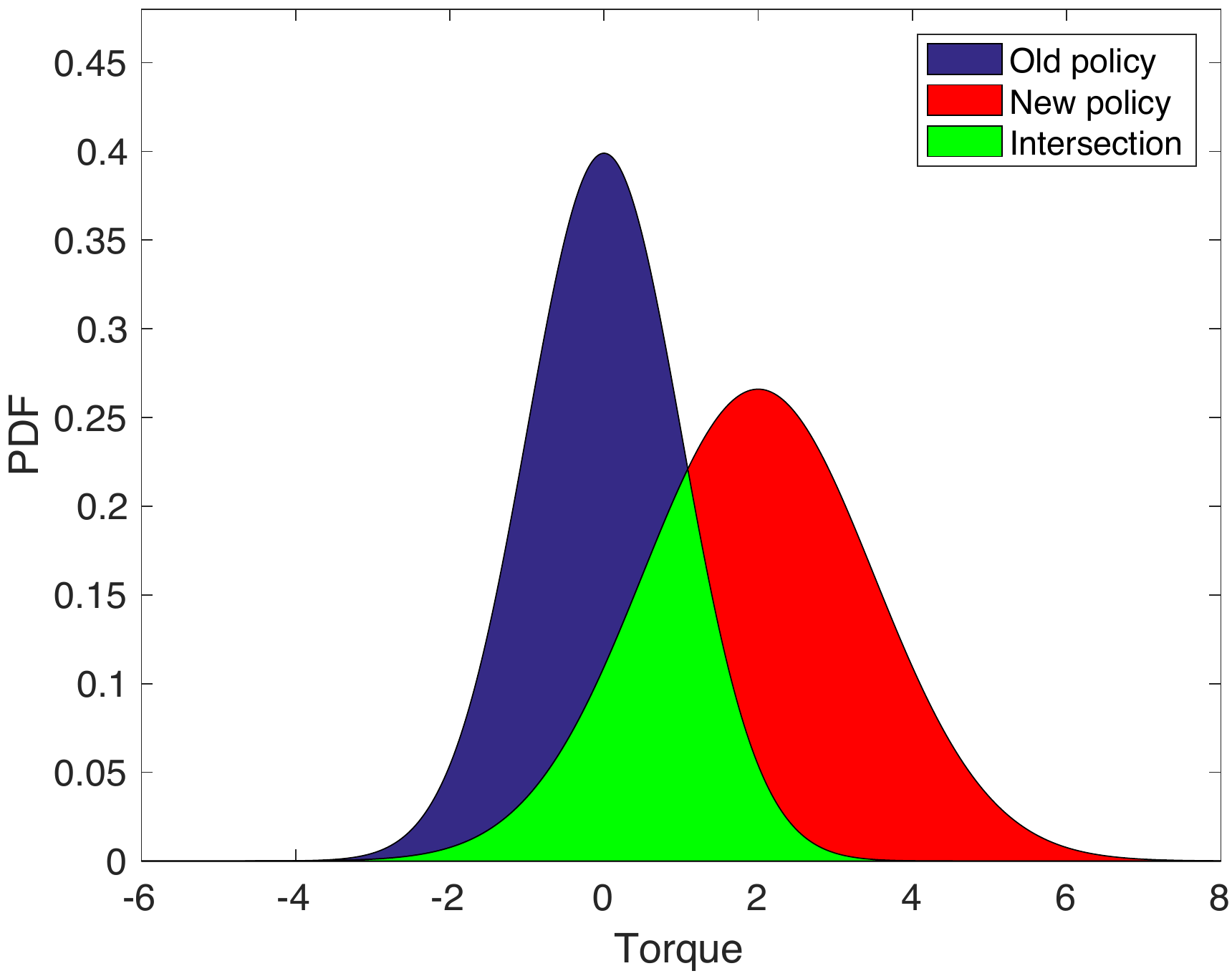}
		\end{center}
		\caption{Our policy is represented by a Gaussian, shown in blue and green.  After updating the policy parameters, the new policy is shown in red and green.  The green area is the probability mass that is consistent between the new and old policies, whereas the blue mass from the old policy was shifted to the red mass of the new policy.  Both distributions integrate to 1, so the blue and red areas must be equal. (Best viewed in color)}
		\label{fig:update_params}
	\end{figure}

	In order to find the optimal policy in the test environment, we must update the policy parameters.  However, updating the policy parameters can increase the unsafety rate, possibly leading to a violation of the damage constraint. To examine the effect of changing the policy parameters on the unsafety rate, we compute the amount that the policy distribution can change.  An example is shown in Figure~\ref{fig:update_params}.  The degree to which the policy remains the same is given by the area of intersection, $A_I$, between the new policy and the old policy (shown in green).  The amount that the policy distribution has changed is then given by $1 - A_I$, which is equal to the area of the blue section in Figure~\ref{fig:update_params}.  This probability mass has shifted to the red area of this figure, which must also have an area of $1 - A_I$ in order for the new policy distribution to integrate to 1.  As in the previous section, if we conservatively assume that changes in the policy distribution all lead to unsafe states, then the unsafety rate might increase by
	\begin{align}
	\label{eq:delta_pf2}
	\Delta p_{u2} = (1 - A_I).
	\end{align}
	This quantity is used by our algorithm in Equations~\ref{eq:pf_adjust} and~\ref{eq:tl_adjust} to determine the new torque limit.
		
	If the policy is represented by a complex function such as a neural network, it is typically difficult to predict how updating the parameters will affect the policy distribution.  However, if we are using Trust Region Policy Optimization (TRPO)~\cite{schulman2015trust} to update the policy parameters, then we will have a good estimate on how the distribution can change between iterations.  In TRPO, we update the policy according to the following optimization:
	\begin{align}
	& \underset{\theta}{\text{maximize}}
	& & \E\Big[\frac{\pi_\theta(a | s)}{\pi_{\theta_{old}}(a | s)}A_{\theta_{old}}(s,a)\Big] \\
	& \text{subject to}
	& & \E[D_{KL}(\pi_{\theta_{old}}(\cdot|s) \, || \, \pi_\theta(\cdot|s))] \leq \delta_{KL}
	\end{align}
	where $\theta_{old}$ are the old policy parameters, $\theta$ are the new policy parameters, $D_{KL}$ is the KL-divergence, $\delta_{KL}$ is a constant, and $A_{\theta_{old}}(s,a)$ is the advantage function, computed as the empirical return minus a baseline~\cite{duan2016benchmarking}.  The constant $\delta_{KL}$ limits the amount by which the KL-divergence of the policy can change on each iteration.  Thus, although we do not know exactly how the policy has changed from one iteration to the next, when using TRPO we know that the KL-divergence of the policy cannot change more than $\delta_{KL}$.  
	
\begin{figure}[htb]
			\begin{center}
				\includegraphics[width=0.6\linewidth]{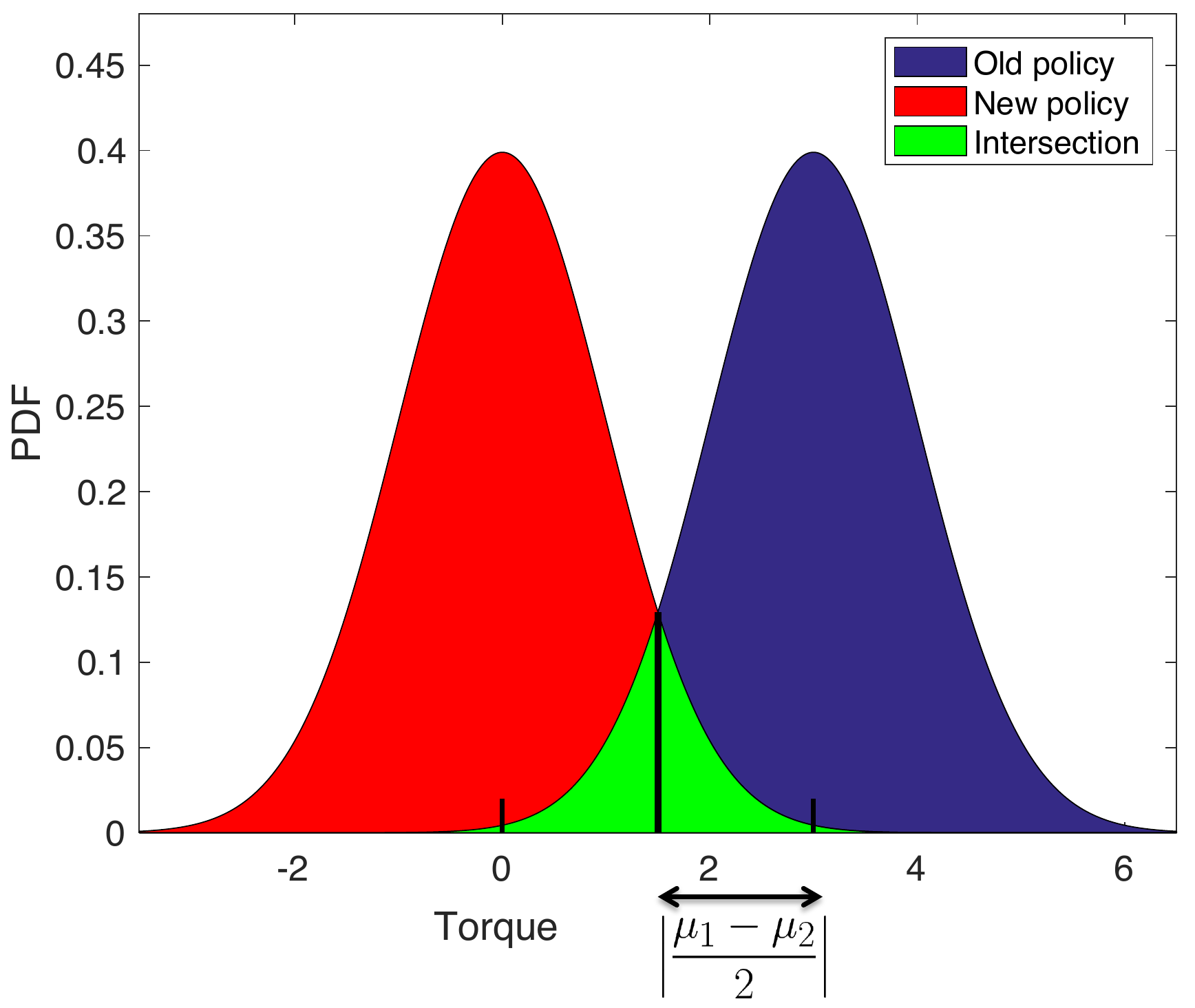}
			\end{center}
			\caption{Our old policy is represented by a Gaussian, shown in blue and green.  After updating the policy parameters, the new policy is shown in red and green.  The variances of the two distributions are assumed to be approximately equal.  (Best viewed in color)}
			\label{fig:update_params_kl}
		\end{figure}

	We would like to use this information to estimate $(1 - A_I)$.  The KL-divergence between two Gaussian distributions is given by
	\begin{align}
	KL(p_1 || p_2) = \log\frac{\sigma_2}{\sigma_1} + \frac{\sigma^2_1 + (\mu_1 -\mu_2)^2}{2\sigma^2_2} - \frac{1}{2}
	\end{align}
	where $p_1=\mathcal{N}(\mu_1, \sigma^2_1)$ and $p_2 = \mathcal{N}(\mu_2, \sigma^2_2)$.  If we make the simplifying assumption that the variance stays roughly the same between each iteration (i.e. $\sigma_1 = \sigma_2 = \sigma$), and we set the KL divergence equal to the constraint $\delta_{KL}$, then we get
	\begin{align}
	|\mu_1 - \mu_2| = \sigma\sqrt{2 \, \delta_{KL}}.
	\end{align}
	As illustrated in Figure~\ref{fig:update_params_kl}, the area of intersection between the two Gaussians of equal variance would then be given by
	\begin{align}
	A_I(\mu_1, \sigma) &= 2F_{\mu_1, \sigma}\Big(\mu_1 - \Big|\frac{\mu_1 - \mu_2}{2}\Big|\Big) \\
	&= 2 \, F_{\mu_1, \sigma}(\mu_1 - \sigma\sqrt{\delta_{KL}/2})
	\end{align}
	where $F_{\mu_1, \sigma}(\cdot)$ is the cumulative distribution function for a normal distribution of mean $\mu_1$ and variance $\sigma^2$.  
	The mean and variance are conditioned on the state, so as before, we can use an expectation to remove the dependency on the state:
	\begin{align}
	\label{eq:ai_exp}
	A_I = 2 \, \E_{s \sim p(s)} [F_{\mu_1, \sigma}(\mu_1 - \sigma\sqrt{\delta_{KL}/2})].
	\end{align}
	We can then use this estimate in Equation~\ref{eq:delta_pf2} to predict how much the unsafety rate might increase as a result of updating our policy.
	
	\section{Results}
	\subsection{Experimental Setup}
\begin{figure}[th]
		\begin{center}
			\includegraphics[width=0.95\linewidth]{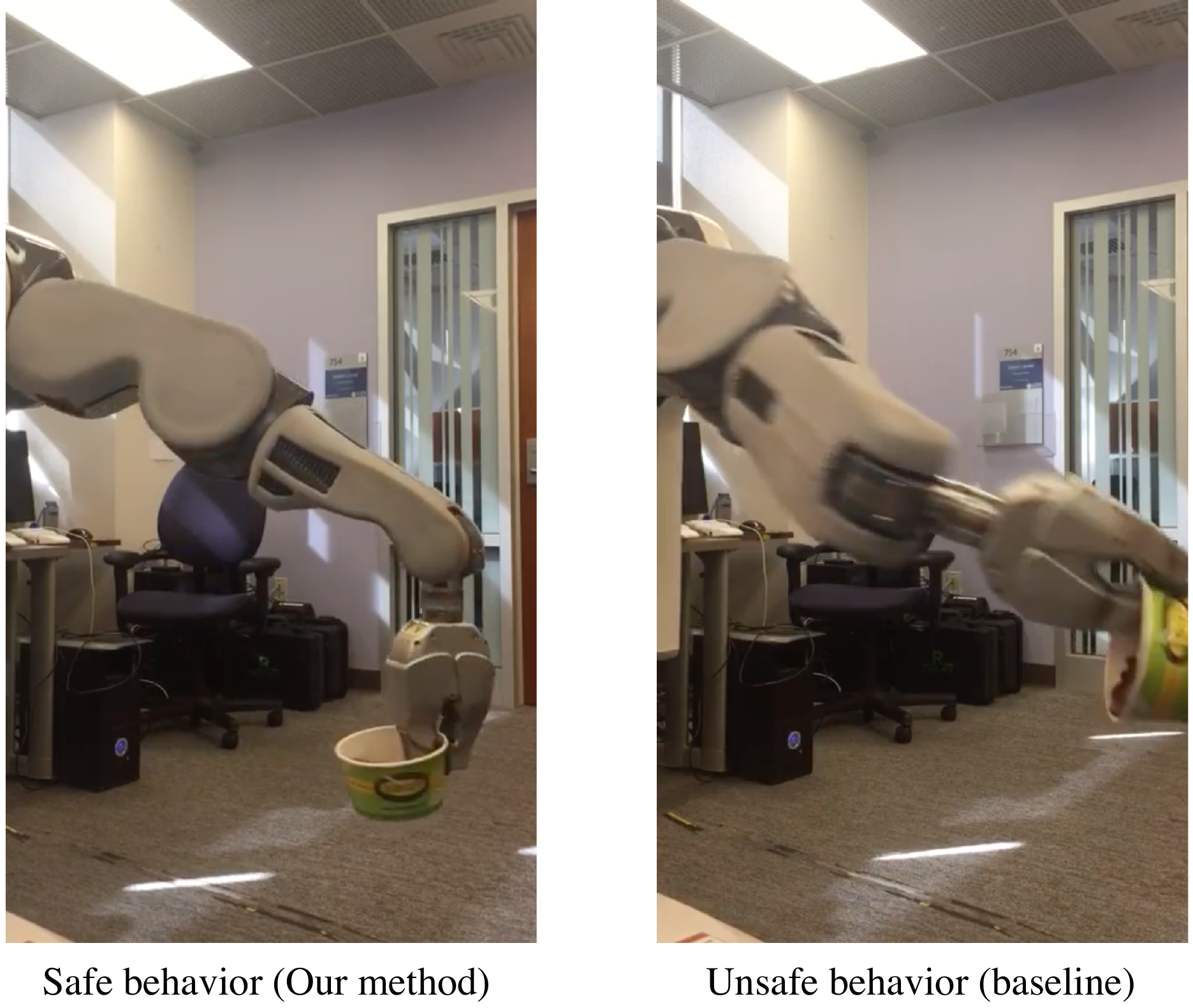}
		\end{center}
		\caption{Safe behavior (our method, left) vs unsafe behavior (baseline, right) for the cup swirl task.  The goal is to swirl a full cup of almonds without spilling.}
		\label{fig:cup_swirl}
	\end{figure}

	We test our method both with a simulated PR2 (Section~\ref{sec:sim_results}) and with a real PR2 (Section~\ref{sec:real results}); pictures of each robot can be seen in Figure~\ref{fig:break}. To illustrate our method, we use as an example task that of swirling a cup in a circle while keeping the cup oriented vertically (so that the contents do not spill).  Examples of safe and unsafe behavior for this task can be seen in Figure~\ref{fig:cup_swirl}, and videos of our results are available online\footnote{\url{https://youtu.be/fprZHyP\_50o}}.
	
	Our safety limit is defined with respect to the orientation of the end-effector; in our experiments, the end-effector must point straight down and maintain a fixed vertical orientation.  If $\theta$ is the angle of the end-effector with respect to the downward direction, then our safety constraint is defined as $\theta \leq \theta_{\rm lim}$, where $\theta_{\rm lim}$ is defined as the maximum angle that the robot can hold the cup without a risk of spilling.  The unsafety rate $p_u$ then measures the fraction of timesteps for which the $\theta > \theta_{\rm lim}$.  We assume that the amount that is spilled (i.e. the damage) will be greater if the constraint is violated at high torques (i.e. if the robot tips the cup and applies a large torque, moving quickly).  Additionally, our experiments demonstrate that, at higher torques, the robot will have a harder time recovering from small errors, leading to larger errors and more damage.


	
	The reward is designed to encourage the robot to swirl the cup in a horizontal circle.  The modified reward used by our reinforcement learning algorithm is defined using the penalty method approach of equation~\ref{eq:penalty_reward}.  In our experiments, we set $u'_{\rm lim} = 0$ and $\lambda = 0.001$.
Our state space is defined by the left arm's joint-angles, the joint velocities, and the vector $z$ which measures the downward orientation of the end-effector; for optimal safety, we should have $z = (0, 0, -1)$.  The actions of our policy are the 7 joint torques for the left arm of the PR2.
	
	
	We pre-train the policy in simulation using the Mujoco~\cite{todorov2012mujoco} simulator.  Our policy is represented as a neural network with 3 hidden layers of 64 nodes each, with tanh non-linearities after each hidden layer.  The policy is optimized using TRPO~\cite{schulman2015trust} and Generalized Advantage Estimation~\cite{schulman2015high} with a linear baseline, as implemented by rllab~\cite{duan2016benchmarking}.  Each episode consists of 200 timesteps, where each timestep is 50 milliseconds, and we pre-train the policy with 50 episodes per batch.  We use a discount factor of $\gamma$ = 0.95, $\lambda = 0.98$ from Generalized Advantage Estimation, and a KL-divergence constraint of $\delta_{KL} = 0.01$.  During pre-training, we re-initialize the robot arm to a random initial configuration before each episode, so that the policy will be robust to the initial position of the robot.

	After pre-training, we test our policy on a robot that may have different model parameters from those that were used during pre-training.  For example, if we are pre-training in simulation and then transferring our policy to the real world, the real robot may have different values for friction, damping, or other model parameters. 
	
	Our policy is fine-tuned in the test environment in order to learn a safe policy that will correctly implement the desired behavior. 	Because we desire for our policy to quickly converge during fine-tuning, we increase our KL-divergence constraint to $\delta_{KL} = 0.05$ and we use only 5 episodes per batch.  Other parameters are kept the same as during pre-training, except as specified below.    In order to ensure that the state-visitation distribution stays relatively fixed from one time-step to the next, we ensure that the torque limit $T_{\rm lim}$ does not increase by more than 5\% between each iteration (for our method as well as for the baselines). This is similar to the KL-divergence constraint that is placed on the policy, which prevents the policy from changing too dramatically in each iteration.
			
	At test time, we will be varying the torque limit $T_{\rm lim}$ to enforce the expected damage limit, using our method described in Section~\ref{sec:Method}.  In order to make the training environment match to the test environment as much as possible, we also vary the torque limit during training.  We additionally input the torque limit into the network as an additional input; thus, our policy can learn a range of behaviors for varying torque limits.  Although this is not necessary for our method to work (our method will still keep the expected damage below the limit even without this modification), our experiments show that this change is helpful for fast convergence of the policy during test time.
		
	\subsection{Simulation Results}
	\label{sec:sim_results}
	Our simulated test environment allows us to test the robustness of our algorithm under controlled environmental variations; we will show results on a real PR2 in Section~\ref{sec:real results}.  We test our algorithm for Probabilistically Safe Policy Transfer in a simulated test environment that differs from our training environment in that 
	all masses and damping parameters are reduced by a factor of 10-100 and moments of inertia are reduced by a factor of 10.
	
	We initialize the torque limit $T_{\rm lim}$ to 0.1 N$\cdot$m and increase it as specified by our method up to a maximum value of 3 N$\cdot$m.
	We use $D_{\rm safe} = 0.5$, except where indicated otherwise.
	For each method that we evaluate below, we run the method with 5 random seeds, in order to test the robustness of each approach.
	
\begin{figure}[ht]
		\begin{center}
			\includegraphics[width=0.85\linewidth]{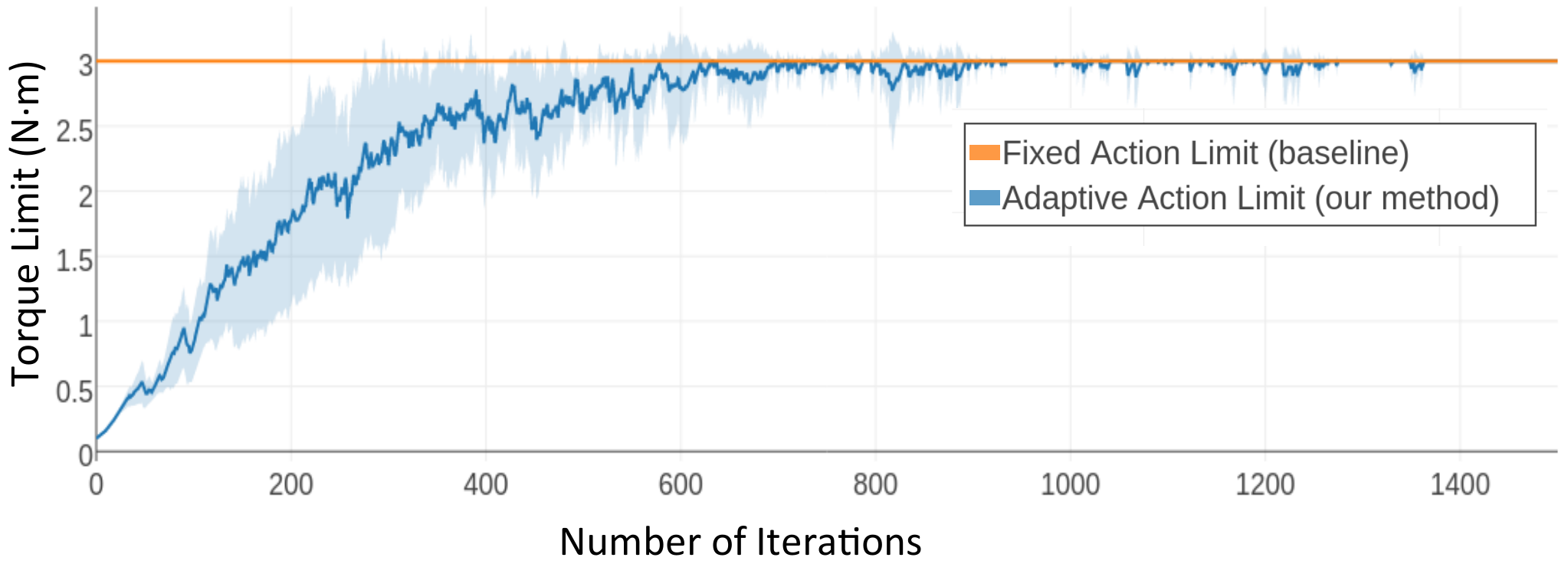}
		\end{center}
		\caption{Our method (shown in blue) adjusts the torque limit over time, based on the safety of the policy.  We compare to a baseline (shown in orange) which has a fixed torque limit, as is commonly done. The lines show the mean and variance of the torque limit over 5 runs with different random seeds.  (Best viewed in color)}
		\label{fig:action_limit}
	\end{figure}

	\textbf{Adaptive vs Fixed Torque Limit.}
	First, we show the benefit of using our adaptive torque limit method compared to a baseline of using a fixed torque limit.  Shown in Figure~\ref{fig:action_limit}, our method starts with a low torque limit (0.1 N$\cdot$m) and adjusts the torque limit over time.  In contrast, the baseline uses a fixed torque limit of 3 N$\cdot$m.  Note that our method does not always increase the torque limit; sometimes it may decrease it based on the unsafety rate.
	
\begin{figure}[ht]
		\begin{center}
			\includegraphics[width=0.85\linewidth]{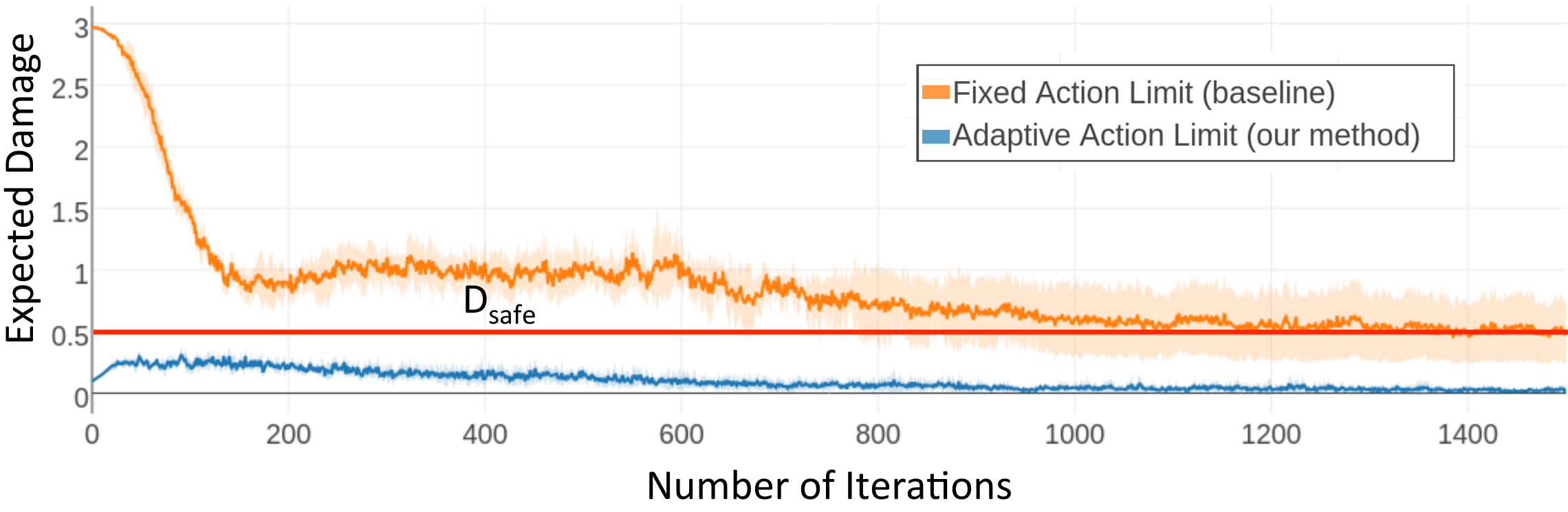}
		\end{center}
		\caption{By adapting the torque limit based on the performance, our method (shown in blue) is able to keep the expected damage below the safety limit (shown in red).  In contrast, using a fixed torque limit leads to a large expected damage, especially during the initial iterations while the policy is still adapting.  The plot shows the mean and variance of the expected damage over 5 runs with different random seeds.   (Best viewed in color)}
		\label{fig:expected_damage}
	\end{figure}

	Figure~\ref{fig:expected_damage} shows the expected damage from each of these approaches.  During the initial iterations of the policy, both methods perform relatively poorly, leading to a large number of constraint violations and hence a high unsafety rate ($p_u$).  The fixed torque limit approach operates at large torques even during these initial stages, leading to a large expected damage ($p_u \, T_{\rm lim}$), as shown.  In contrast, our adaptive torque limit approach initially operates at a low torque limit (0.1 N$\cdot$m), and thus the initial expected damage is kept low.  Our method adapts the torque limit appropriately to keep the expected damage below $D_{\rm safe}$ as the policy adapts.
	
	\textbf{Ablation Analysis.}
	Next, we perform an ablation analysis to demonstrate the importance of each component of our method.  In our full method, we predict that the probability of unsafety $p_u$ might increase from its current value due to two factors:
	\begin{itemize}
		\item Changing the torque limit might cause $p_u$ to increase by $\Delta p_{u1}$ (Section~\ref{sec:Varying Torque Limit})
		\item Updating the policy might cause $p_u$ to increase by $\Delta p_{u2}$ (Section~\ref{sec:Updating the params})
	\end{itemize}
	Thus, our full method predicts that, as a result of both of these factors, the probability of unsafety $p_u$ might increase to
	\begin{align}
	p'_{u} = p_u + \Delta p_{u1} + \Delta p_{u2}
	\end{align}
	We then set the torque limit to a value of $T_{\rm lim} = D_{\rm safe} / p'_u$, ensuring that the expected damage constraint will continue to be satisfied.
		
	In this section, we explore the effect of removing different components of our method; specifically, we explore the following variants for the predicted probability of unsafety:
	\begin{enumerate}[label=v\arabic*)]
		\setcounter{enumi}{1}
		\item Not predicting effect of the torque limit increase: \\ $p'_{u} = p_u + \Delta p_{u2}$ 
		\item Not predicting effect of the policy update: \\ $p'_{u} = p_u + \Delta p_{u1}$ 
		\item Not predicting either effect: \\ $p'_{u} = p_u$
	\end{enumerate}
	In each case, we set the torque limit to a value of $T_{\rm lim} = D_{\rm safe} / p'_u$ as before.  We run each method with 5 random seeds.
	
\begin{figure}[ht]
		\begin{center}
			\includegraphics[width=0.74\linewidth]{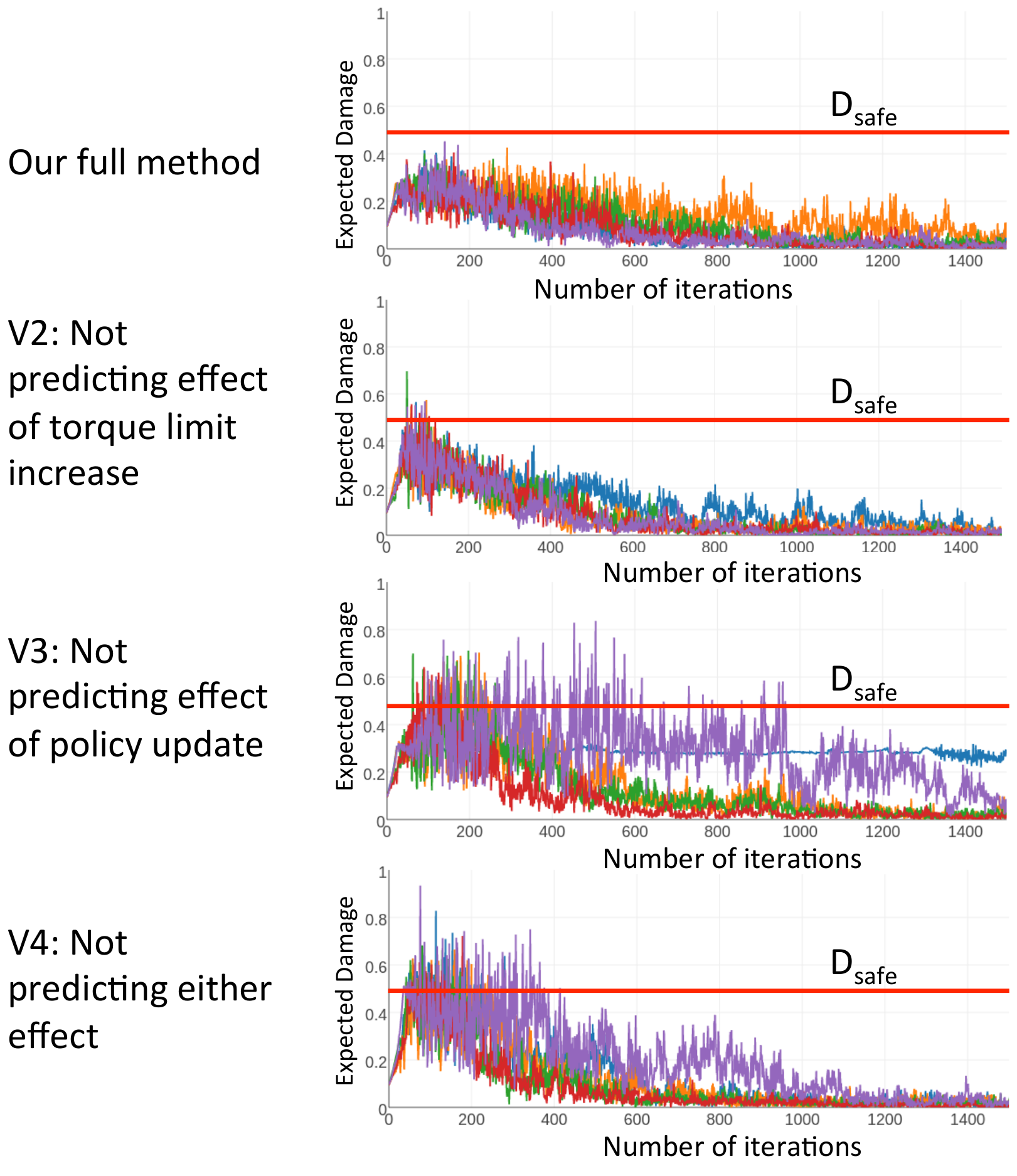}
		\end{center}
		\caption{Ablation analysis: We compare our full method (top) to various modified versions of our method, removing different components of our approach (rows 2 through 4).  Our full method is the only version that does not violate the expected damage limit $D_{\rm safe}$.}
		\label{fig:ablation}
	\end{figure}

	The result of our ablation analysis is shown in Figure~\ref{fig:ablation}.  As can be seen in this figure, removing any part of our method results in a violation of the expected damage limit $D_{\rm safe}$.  Thus, in order to observe the expected damage limit, it is important to predict both how changing the torque limit and how updating the policy parameters will affect the probability of unsafety.
		
\begin{figure}[ht]
		\begin{center}
			\includegraphics[width=0.65\linewidth]{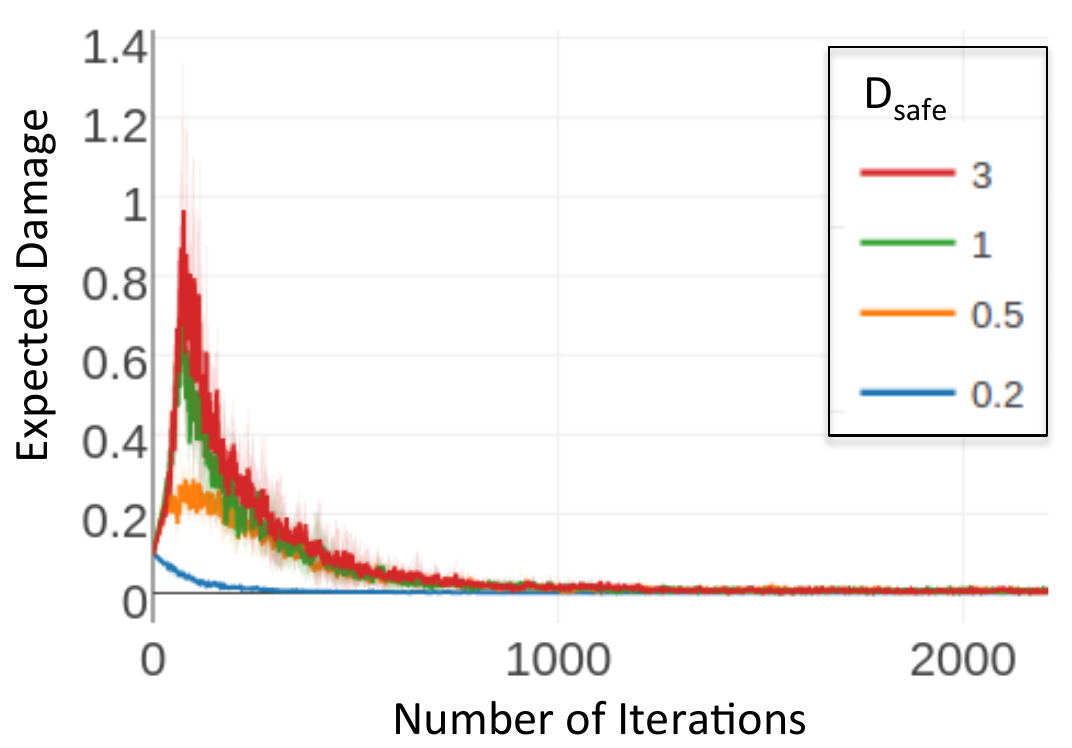}
		\end{center}
		\caption{We show the effect on our algorithm of varying the expected damage limit, $D_{\rm safe}$.  Each line represents the expected damage for a different value of $D_{\rm safe}$.  The goal of our method is to ensure that, in each case, the expected damage limit stays below $D_{\rm safe}$. (Best viewed in color)}
		\label{fig:d_safe}
	\end{figure}

	\textbf{Varying the expected damage limit.}
	In Figure~\ref{fig:d_safe}, we show the effect on our policy of varying $D_{\rm safe}$, the expected damage limit.  We test our method with four different values of $D_{\rm safe}$.  In each case, our method adjusts the torque limit accordingly to try to keep the expected damage below $D_{\rm safe}$.  In some trials we observed the emperical expected damage exceed the safety threshold due to the stochasticity of our policy in small batch sizes; with larger batch sizes, our method observed the safety limit in all of our experiments.
	
	\subsection{Real-world Results}
	\label{sec:real results}
\begin{figure}[ht]
			\begin{center}
				\includegraphics[width=0.95\linewidth]{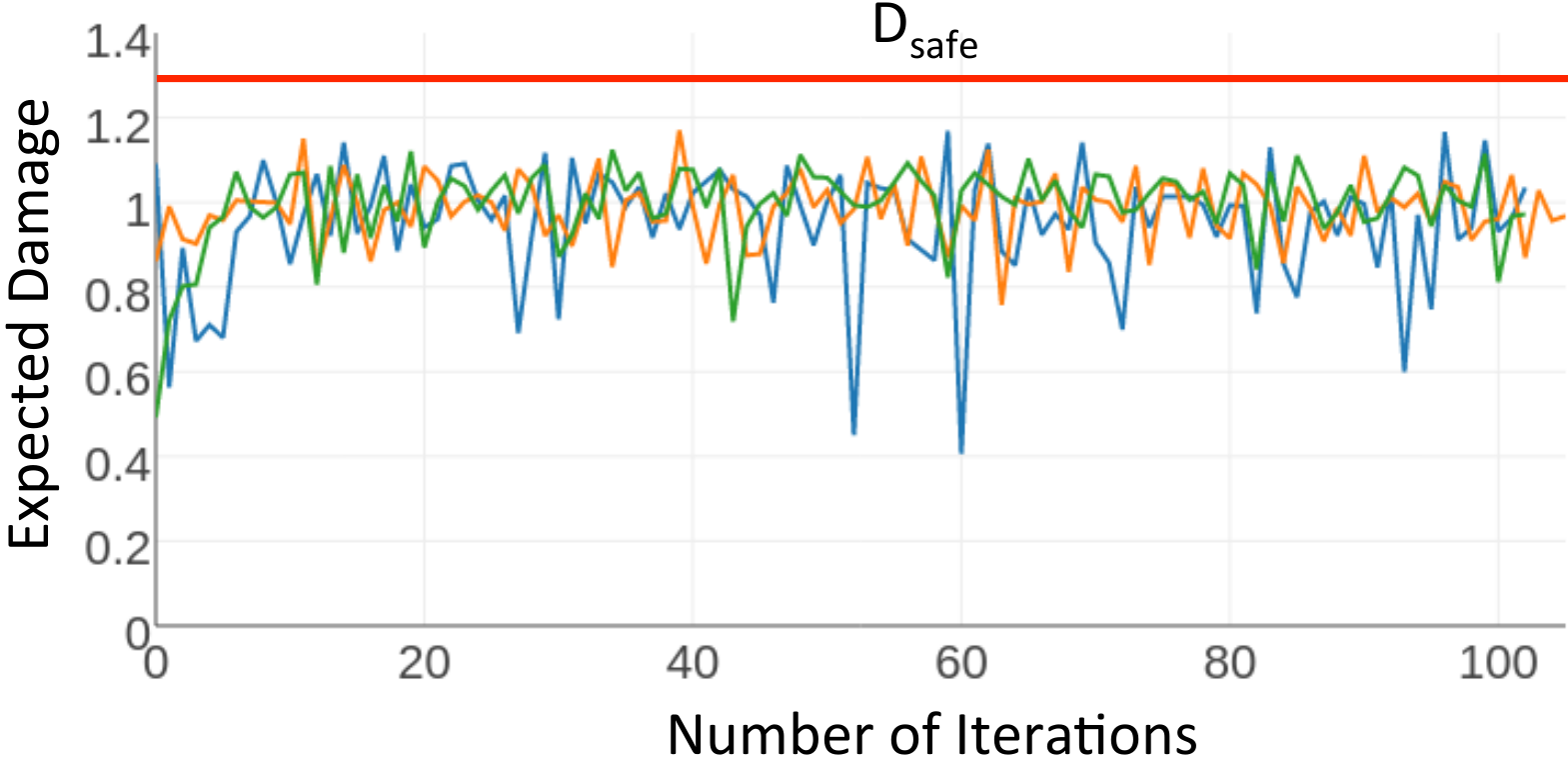}
			\end{center}
			\caption{By adapting the torque limit based on the performance, our method is able to keep the expected damage below the limit $D_{\rm safe}$. Each line shows our method with a different random seed.}
			\label{fig:expected damage real_swirl}
		\end{figure}

	To test whether our method works in the real world, we also fine-tune our pre-trained policy using the real PR2.  Our real robot appeared to have much greater friction than that of our simulation; to overcome the static friction, we initialized the torque limit $T_{\rm lim}$ to 1.8 N$\cdot$m, and we increased all torques output by the policy by 30\%.  We set $D_{\rm safe} = 1.3$.  To avoid unsafe actions due to the stochasticity of the policy, we reduced the policy variance by a factor of two for real-world testing.
	We run our method three times, with different random seeds.  As shown in Figure~\ref{fig:expected damage real_swirl}, our method succeeds in keeping the expected damage below the damage limit.  
	
	We also tested whether our method can be used to successfully train a robot to swirl a full cup without spilling.  In order to verify this, we filled a cup with almonds and tested both our method as well as the baseline (with no torque limits).  For each method, we counted the number of almonds that were spilled after 10,000 timesteps of our policy (8.3 minutes).  Based on 5 repeated experiments, our method drops an average of 7.8 almonds, compared to 18.2 almonds for the baseline (this result is statistically significant with $p < 0.05$).  Videos of our results are available online\footnote{\url{https://youtu.be/fprZHyP\_50o}}

	
	\section{Conclusion}
	We present a probabilistic framework for defining robot safety in terms of maximizing the expected return while keeping the expected damage below a given limit.  As the policy adapts to its new environment, our method adjust the torque limit accordingly to ensure that the robot remains safe.  Importantly, our method predicts how changes to the policy or changes to the torque limit might affect the unsafety rate, in order to keep the expected damage below the limit at every iteration.
	
	\section{Acknowledgments}
	This work was partially supported by an NSF CAREER award (\#1351028)
	and by Berkeley Deep Drive.
	
	\bibliographystyle{splncs03}
	\bibliography{safetuning}

\end{document}